\pdfoutput=1

\documentclass[11pt]{article}

\usepackage{acl}

\usepackage{times}
\usepackage{latexsym}
\usepackage{array}
\usepackage{multirow}
\usepackage{graphicx} 

\usepackage{multirow}
\usepackage{kotex}
\usepackage{multicol}
\usepackage{float}
\usepackage{enumitem} 
\usepackage[most]{tcolorbox} 
\usepackage{caption} 

\usepackage[T1]{fontenc}

\usepackage[utf8]{inputenc}

\usepackage{microtype}

\usepackage{inconsolata}

\title{Efficient Terminology Integration for LLM-based Translation in Specialized Domains}

\author{
Sejoon Kim, 
Mingi Sung, 
Jeonghwan Lee, 
Hyunkuk Lim, 
Jorge Froilan Gimenez Perez \\
PwC Korea GenAI Team, Seoul, South Korea \\
\textit{\{sejoon.s.kim, mingi.sung, jeonghwan.lee, hyunkuk.lim, gimenez.jorge\}@pwc.com} \\
}

\begin{document}
\maketitle
\begin{abstract}
Traditional machine translation methods typically involve training models directly on large parallel corpora, with limited emphasis on specialized terminology. However, In specialized fields such as patent, finance, or biomedical domains, terminology is crucial for translation, with many terms that needs to be translated following agreed-upon conventions. In this paper we introduce a methodology that efficiently trains models with a smaller amount of data while preserving the accuracy of terminology translation. We achieve this through a systematic process of term extraction and glossary creation using the Trie Tree algorithm, followed by data reconstruction to teach the LLM how to integrate these specialized terms. This methodology enhances the model's ability to handle specialized terminology and ensures high-quality translations, particularly in fields where term consistency is crucial. Our approach has demonstrated exceptional performance, achieving the highest translation score among participants in the WMT patent task to date, showcasing its effectiveness and broad applicability in specialized translation domains where general methods often fall short.
\end{abstract}

\section{Introduction}

\begin{figure}[h]
    \centering
    \resizebox{0.5\textwidth}{0.2\textheight}{\includegraphics{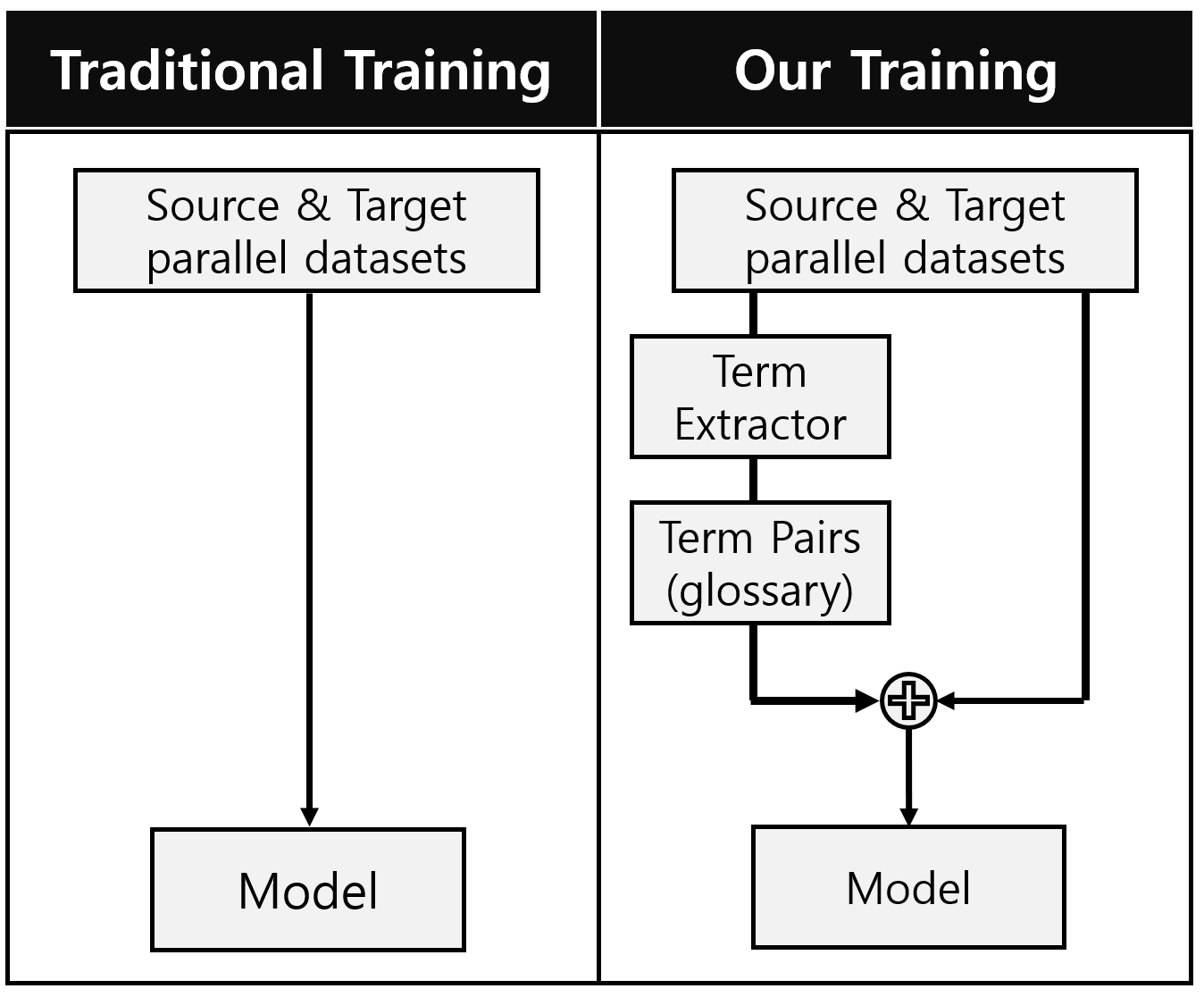}}  
    \setlength{\abovecaptionskip}{0pt}
    \setlength{\belowcaptionskip}{0pt}
    \caption{Training method in terminology-based LLM translation}
    \label{fig:training_diff}
\end{figure}

Conventional approaches to machine translation typically rely on training models using extensive parallel corpora, with little focus on specialized vocabulary. While this can be an effective approach in general, it demands large amounts of data and may lead to inconsistent translations of technical or domain-specific terminology. This challenge is particularly acute in specialized fields, where precise terminology usage is crucial and high-quality training data is often scarce~\cite{skianis-etal-2020-evaluation, ghazvininejad2023dictionarybasedphraselevelpromptinglarge, zhang-etal-2023-understanding}. Datasets for training models in these specialized domains are usually limited, and even when they exist, many are private due to security concerns. Consequently, certain industries lag behind in the advancement of deep learning-based translation. This disparity is even more pronounced for less commonly spoken languages, where specialized translation capabilities are significantly underdeveloped, resulting in an unequal distribution of progress in neural machine translation.

Numerous approaches have been explored to integrate terminology constraints into Neural Machine Translation (NMT) systems, aiming to improve domain-specific translation quality. Recent research on terminology-based machine translation has shifted towards incorporating constraints during the training phase, which eliminates the computational overhead during inference and enhances translation quality. \citet{DBLP:journals/corr/abs-1906-01105} introduced a method where NMT models are trained with augmented datasets that include terminology constraints as inline annotations, allowing the model to learn the appropriate use of these terms during training. Building on this, \citet{ailem-etal-2021-encouraging} proposed further enhancements by using token masking and a modified cross-entropy loss function, which biases the model towards generating constraint terms more effectively. Additionally, the use of large language models for post-translation refinement has been explored to improve terminology recall, demonstrating the evolving nature of terminology integration in NMT~\cite{bogoychev-chen:2023:WMT, ghazvininejad2023dictionarybasedphraselevelpromptinglarge, moslem-EtAl:2023:WMT}. These training-based approaches have demonstrated significant improvements in both BLEU scores and terminology usage rates compared to decoding-time methods, indicating their effectiveness in satisfying lexical constraints without compromising translation quality.

In this paper, we propose a fine-tuning approach to resolve the domain-specific terminology mismatch problem using only a small dataset. Our approach focuses on extracting a glossary from the existing training datasets and fine-tuning the model to integrate these terms effectively into translations. First, we train a terminology extraction model to generate a glossary from existing training datasets, which we integrate into our trie data structure~\cite{BODON2003739}. We then extract domain-specific terms from the source sentences using the tree structure and pass them along with the source texts to instruct our Large Language Model (LLM) to effectively incorporate specialized terminology into translations. This approach ensures high-quality and consistent results in specialized fields. Figure~\ref{fig:training_diff} illustrates how our approach differs from traditional fine-tuning methods. This targeted refinement process enhances the model's capacity to manage specialized terminology, thereby maximizing the utility of the original training data and significantly improving translation accuracy and consistency. Our methodology has proven to be exceptionally effective, particularly in specialized translation tasks, where general translation methods often struggle to maintain accuracy and consistency. Notably, our approach achieved the highest translation score among all participants in the WMT patent task, underscoring its superior performance and broad applicability across various specialized translation domains. Through this systematic and targeted strategy, we ensure that our translations are not only accurate but also contextually relevant, thereby providing a reliable solution for specialized translation needs.

\section{Methodology}
In this section, we describe the methodology employed in developing a domain-specific terminology-based LLM translation system, focusing on three key processes: (1) the creation of a terminology glossary, (2) the identification of terms within the source text, and (3) the application of these terms during the translation process using LLM prompts and sLLM fine-tuning.

\subsection{Construction of the Terminology Glossary: Terminology Aligner}

\begin{figure}[htbp]
\centering
\newtcolorbox[auto counter, number within=section]{breakablebox}[2][]{colback=white, colframe=black, coltitle=black, boxrule=0.8mm, sharp corners, title=#2, #1, breakable}

\begin{breakablebox}[label={sec:instructions}]{Instructions for Term Extraction}

\textbf{System Message:}

I will now show you source sentences in Japanese and target sentences in Korean. Your task is to extract and pair key terms from both the original and translation texts. Maintain the exact form of the terms without modification.

Please follow these instructions for extracting term pairs:

\begin{itemize}[noitemsep, topsep=0pt, partopsep=0pt, parsep=0pt] 
    \item Extract term pairs that are closely related to patents.
    \item Only extract nouns.
    \item The extracted term pairs will be used to create a Japanese-Korean glossary.
    \item Return the results in the form of a Python dictionary, as shown in the example.
    \item However, if the exact same term appears more than once include it only once.
\end{itemize}
\vspace{\baselineskip} 
\small
\textbf{Example 1:}

\begin{verbatim}
src_sentence = それぞれについて官能評価を行った結果を表４２に示す。

tgt_sentence = 각각에 대하여 관능 평가를 행한 결과를 표 42에 나타낸다。

result = {"官能評価": "관능 평가"}
\end{verbatim}

\textbf{Example 2:}

\small
\begin{verbatim}

src_sentence = 各種の特許権や技術標準化に関する問題が検討された。

tgt_sentence = 각종 특허권과 기술 표준화에 관한 문제가 검토되었다。

result = {"特許権": "특허권", "技術標準化": "기술 표준화"}
\end{verbatim}

\end{breakablebox}
\setlength{\abovecaptionskip}{0pt}
\setlength{\belowcaptionskip}{0pt}
\caption{Instructions for Term Extraction}
\label{fig:aligner_instructions}
\end{figure}

To enable the translation model to produce accurate translations that incorporate specialized terminology, we first construct a "Terminology Pair Dictionary," aligning key terms between the source and target languages. We achieve this by fine-tuning the Mistral Nemo model, creating a Terminology Aligner model whose primary objective is to extract pairs of key terms from both the original and translated texts.

For our training data, we leverage the GPT-3.5 API to generate synthetic data by crafting prompts that instruct the API to extract key term pairs from existing Japanese-Korean translation pairs in our dataset, along with the system prompt shown in Figure~\ref{fig:aligner_instructions}. From the 1,000,000 training samples provided by the organizers, we randomly select 1,000 examples to fine-tune Mistral Nemo for a single epoch. We adopt this conservative approach, recognizing that Mistral Nemo already possesses a robust grasp of both Korean and Japanese and is capable of performing various tasks, including the one at hand. Our goal is to specialize the model for our particular task without compromising its broader capabilities or confusing it with unrelated tasks.

Furthermore, when the entire dataset was used for fine-tuning, the model frequently extracts non-essential term pairs or entire sentences as pairs, indicating overfitting. By carefully selecting the amount of data and limiting the number of training epochs, we ensure that it extracts only the most relevant, domain-specific term pairs and effectively fine-tune the Terminology Aligner model.

\subsection{Term Identification in the Source Text: Trie-Tree Algorithm}

\begin{figure}[h]
    \centering
    \resizebox{0.5\textwidth}{0.15\textheight}{\includegraphics{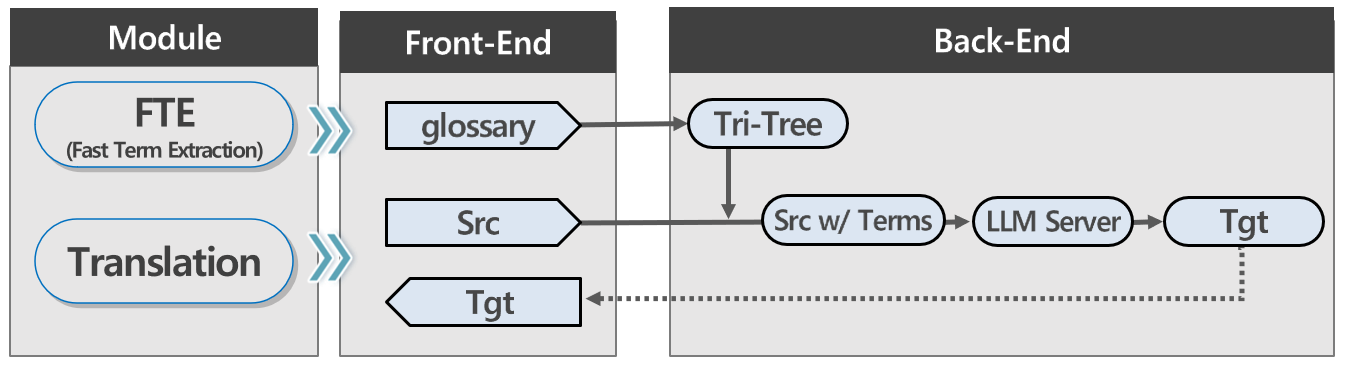}}  
    \setlength{\abovecaptionskip}{0pt}
    \setlength{\belowcaptionskip}{0pt}
    \caption{Overall process of term extraction to translation}
    \label{fig:trie_tree_example}
\end{figure}

The next step in our methodology involves identifying and extracting specialized terms from the source text that must be accurately translated using the glossary we constructed. To account for industries where there is often a high volume of technical terms and the need for efficient text scanning, we implement the Trie Tree data structure to extract the domain-specific terms.

The Trie Tree is particularly well-suited for this task due to its efficiency in string searching and matching. The algorithm operates by placing a cursor at the first Unicode character of the text, while another cursor points to the root of the tree. As the text cursor advances through each character, the tree cursor checks for corresponding child nodes. If a match is found, the tree cursor moves to the next node; if not, it resets to the root. When the cursor reaches a node marked as a 'term,' the term is identified, and its position is recorded. This allows us to quickly retrieve the term's translation and include it in the LLM prompt, ensuring that all relevant terms in the text are accurately and efficiently identified. The process is visually illustrated in Figure~\ref{fig:trie_tree_example}, which describes the step-by-step progression of the Trie Tree algorithm from text scanning to term retrieval and integration into the LLM prompt.

\subsection{Application of Terms in Translation: LLM Prompting and sLLM Fine-Tuning}

\newtcolorbox[auto counter, number within=section]{breakablebox2}[2][]{colback=white, colframe=black, coltitle=black, boxrule=0.8mm, sharp corners, title=#2, #1, breakable}

\begin{figure}[htbp]
\centering
\begin{breakablebox2}[label={sec:instructions}]{Instructions for Term Extraction}

\textbf{System Message:}

You are a professional translator. You are especially familiar with specialized patent knowledge and terms in chemistry, electricity, mechanical engineering, and physics, as well as general everyday terms. Translate the following Japanese source text into Korean.

\begin{itemize}[noitemsep, topsep=0pt, partopsep=0pt, parsep=0pt]
    \item Refer to the word pairs in the glossary when you translate.
    \item Do not translate the glossary itself.
    \item Do not include anything but translation result only.
    \item If a term in the glossary has multiple possible translations separated by '|', choose the most appropriate one.
    \item The translation result must be written in a single line. There must be no newline character at the end.
\end{itemize}

\vspace{\baselineskip}
\textbf{Glossary:}
\small
\begin{verbatim}
{セレノール化合物 :  셀레놀 화합물,
端部 : 끝부분 | 단부 | 모서리 ,
絶縁膜 : 절연막,
送信回路 : 송신 회로 | 전송 회로}
\end{verbatim}
\end{breakablebox2}
\setlength{\abovecaptionskip}{0pt}
\setlength{\belowcaptionskip}{0pt}
\caption{Instructions for Term Extraction}
\label{fig:term_instructions}
\end{figure}

The final phase of our methodology involves the use of the extracted terminology during the translation process. To do this, we first extract term pairs from all translation pairs in our dataset using the created tree structure. These extracted term pairs are then combined with each original translation pair and the system message in Figure~\ref{fig:term_instructions} to create an instruction-based training dataset to fine-tune our translation model.

Similar to our fine-tuning process with the Terminology Aligner, we observed that both the amount of data and the number of training epochs significantly influence the quality of the translation output, particularly in terms of how natural the translations sound. Interestingly, when working with smaller datasets, the model tends to produce more natural, conversational translations. However, as the dataset size increases, the model increasingly adheres to the original sentence structure, resulting in a more formal and literal style of translation.

To balance these tendencies, we use approximately 1,000 data points for training and limit the training to three epochs, with a temperature setting of 0.1. This configuration allows the model to generate translations that were both accurate and natural, making an effective use of the specialized terminology while maintaining a high level of fluency and readability.

\section{Experimental Results and Application}

\begin{table}[h!]
\centering
\normalsize 
\begin{tabular}{|p{0.1cm}|l|c|c|}
\hline
\textbf{} & \textbf{Team} & \textbf{BLEU} & \textbf{RIBES} \\ 
 &  & \textbf{(mecab)} &  \\ \hline
1 & GenAI & \textbf{70.60} & \textbf{0.939073} \\ \hline
2 & Chatgpt (w/ glossary) & 69.00 & 0.929945 \\ \hline
3 & sakura & 68.00 & 0.926839 \\ \hline
4 & Bering Lab & 66.25 & 0.925226 \\ \hline
5 & ryan & 65.74 & 0.922837 \\ \hline
6 & goku20 & 64.30 & 0.922486 \\ \hline
7 & ORGANIZER & 62.43 & 0.915266 \\ \hline
8 & tpt\_wat & 61.00 & 0.918436 \\ \hline
9 & Chatgpt (w/o glossary) & 59.90 & 0.908637 \\ \hline
\end{tabular}
\caption{BLEU (mecab) and RIBES scores for the Japanese-to-Korean translation task}
\label{tab:jk_translation_results}
\end{table}

\begin{table*}[h!]
\centering
\resizebox{\textwidth}{!}{%
\begin{tabular}{|c|l|c|c|c|c|c|c|}
\hline
\textbf{Rank} & \textbf{Team} & \multicolumn{3}{c|}{\textbf{BLEU}} & \multicolumn{3}{c|}{\textbf{RIBES}} \\ \cline{3-8}
 &  & \textbf{juman} & \textbf{kytea} & \textbf{mecab} & \textbf{juman} & \textbf{kytea} & \textbf{mecab} \\ \hline
1 & \textbf{GenAI} & \textbf{67.00} & \textbf{67.40} & \textbf{66.90} & \textbf{0.924474} & \textbf{0.919657} & \textbf{0.923416} \\ \hline
2 & Chatgpt (w/ glossary) & 62.20 & 62.50 & 61.90 & 0.916385 & 0.912133 & 0.914275 \\ \hline
3 & Chatgpt (w/o glossary) & 61.60 & 62.50 & 61.50 & 0.912482 & 0.907932 & 0.911476 \\ \hline
4 & EHR & 53.83 & 55.83 & 54.23 & 0.907358 & 0.903857 & 0.905654 \\ \hline
5 & sarah & 53.59 & 55.68 & 53.94 & 0.903211 & 0.900313 & 0.902430 \\ \hline
6 & KNU\_Hyundai & 53.56 & 55.68 & 54.02 & 0.901627 & 0.900091 & 0.901877 \\ \hline
7 & TMU & 52.85 & 54.92 & 53.24 & 0.906113 & 0.903179 & 0.906320 \\ \hline
8 & Bering Lab & 52.74 & 54.55 & 53.15 & 0.902984 & 0.898627 & 0.902621 \\ \hline
9 & ORGANIZER & 52.02 & 53.93 & 51.99 & 0.897348 & 0.896897 & 0.898316 \\ \hline
10 & sakura & 51.90 & 54.10 & 52.30 & 0.899781 & 0.896489 & 0.898412 \\ \hline
\end{tabular}%
}
\caption{BLEU and RIBES scores for the Korean-to-Japanese translation task}
\label{tab:kj_translation_results}
\end{table*}

Our proposed methodology has been rigorously tested and evaluated within the framework of the WMT patent task, where it achieves the highest translation score to date among all participants. This success demonstrates the effectiveness of our approach in handling domain-specific translations, particularly in maintaining consistency in terminology.

In addition to the translation results generated by our model, we submitted two additional translations using ChatGPT. The first result, labeled 'ChatGPT (w/ glossary)' in Tables 1 and 2, was obtained by replacing our model with ChatGPT while keeping the system prompt and glossary identical to our methodology. The second result was generated using ChatGPT alone without any additional inputs.

Several interesting findings emerged: for the Japanese-to-Korean translation task, ChatGPT without the glossary scores lower than other models in the patent translation domain. However, the score significantly improves when our glossary is provided. This demonstrates that the integration of a terminology glossary substantially enhances translation performance, regardless of the underlying model's capabilities. By comparing ChatGPT with and without the glossary, it becomes evident that our system effectively boosts translation quality through efficient terminology integration. Our specialized language model, trained specifically to use the glossary, outperforms ChatGPT even with the glossary. Upon reviewing the outputs, we notice that ChatGPT sometimes fails to correctly apply terms inside the glossay and occasionally uses Japanese terms instead of their Korean equivalents in the Japanese-to-Korean translation.

These findings highlight that our model can be effectively trained with a small dataset, achieving high-quality translations while remaining a smaller, more efficient model. Beyond patent translation, our methodology can be extended to specialized fields such as legal and financial translation where accurate term alignment is critical, providing a robust solution where general translation methods may fall short.

\section{Discussion}

\subsection{Advantages of Our Methodology Over Traditional Approaches}

The effectiveness of our methodology is further underscored by several key advantages it holds over traditional approaches:

\subsubsection{Focused Learning on Domain-Specific Terms}
Traditional models typically assign equal importance to all words in the training data, which can result in inconsistent translations of specialized terms across different contexts. Our methodology addresses this by prioritizing domain-specific terms, ensuring they are recognized and used consistently in relevant translations.

\subsubsection{Efficient Data Utilization through Terminology Extraction}
Traditional methods often require large volumes of data to achieve satisfactory performance, particularly in specialized domains. Our method optimizes the use of training data by focusing on key term pairs and creating a dedicated glossary, enabling more efficient learning even with a smaller dataset.

\subsubsection{Enhanced Translation Consistency and Accuracy}
A common challenge with traditional translation methods is inconsistency in translating specialized terms, especially when these terms have multiple possible translations depending on context. Our approach mitigates this by ensuring the model is trained with a consistent set of term translations derived from the glossary.

\subsubsection{Improved Model Generalization}
Traditional models trained on large corpora may overfit to specific sentence structures or styles present in the training data, leading to poor generalization to new texts. Our approach incorporates the glossary into training, acting as a regularizing factor that improves generalization to new texts within the same domain.

\subsubsection{Customizability for Different Domains}
Our methodology allows for greater flexibility in adapting the model to different specialized fields. By updating the glossary with new terms relevant to a particular domain, the model can be quickly tailored to perform well without extensive retraining.

\section{Conclusion}

Our terminology-based LLM translation methodology represents a significant advancement in the field of machine translation, particularly for specialized domains requiring precise and consistent term usage. By constructing a terminology glossary using the Terminology Aligner, implementing an efficient term identification process with a Trie Tree algorithm, and fine-tuning the translation process using LLM prompts, we present a system that not only improves translation accuracy but also maintains a high level of naturalness in the output. Our approach has proven successful in terms of performance, operational cost, and training data efficiency, showing great promise for a wide range of professional translation applications.

\bibliography{acl_latex}

\clearpage
\appendix

\section{Appendix}
\label{sec:appendix}

In this appendix, we provide additional details on the training procedures, model configurations, and methodologies employed in our approach for efficient terminology integration in LLM-based translation within specialized domains.

\subsection{Training Details and Additional Information}

For both the translation task and the terminology extraction task, we used the \textbf{Mistral-Nemo-Instruct-2407} model as our base language model. This model was selected due to its strong capability in following instructions, including tasks such as translation and terminology extraction.

\subsubsection{Training Details}

\textbf{Model Configuration and Training Parameters}

{\small
\begin{center}
\begin{tabular}{p{0.45\linewidth} >{\centering\arraybackslash}m{0.45\linewidth}}
\hline
\textbf{Base Model} & \href{https://huggingface.co/mistralai/Mistral-Nemo-Instruct-2407}{mistralai/Mistral-Nemo-Instruct-2407} \\

\multicolumn{2}{l}{\textbf{LoRA Adapter Settings}} \\
\quad Alpha & 8 \\
\quad Rank & 8 \\
\quad Dropout Rate & 0.1 \\
\quad Target Modules & \texttt{["q\_proj", "k\_proj", "v\_proj", "o\_proj", "gate\_proj", "down\_proj", "up\_proj"]} \\

\textbf{Learning Rate} & 1e-5 \\
\textbf{Optimizer} & AdamW  \\
\textbf{Learning Rate Scheduler} & Linear \\
\textbf{Warmup Ratio} & 0.01 \\
\textbf{Epochs} & 1 \\
\textbf{Batch Size} & 4 \\
\textbf{Gradient Checkpointing} & Enabled \\
\hline
\end{tabular}
\end{center}
}

LoRA was applied to mitigate GPU memory limitations and prevent catastrophic forgetting.

\subsection{Fine-Tuning Challenges and Considerations}

Our preliminary experiments indicated that using smaller datasets for fine-tuning resulted in more effective performance for both the terminology aligner and the translation model. Based on these observations, we concluded that a smaller dataset was sufficient to format the model's outputs appropriately and guide it to produce task-specific responses without deviating from the desired content.

The Mistral-Nemo model already exhibited strong abilities in instruction following, including translation and terminology extraction. Therefore, extensive fine-tuning was unnecessary and could potentially degrade performance. Training with larger datasets led to overfitting, where the model's training loss decreased, but the actual translation quality did not improve. In some cases, the model exhibited issues like repetitive outputs. We attempted to mitigate overfitting by increasing dropout rates and weight decay. However, these adjustments did not yield significant improvements in our experiments.

\end{document}